\documentclass[sigconf, nonacm]{acmart}
\AtBeginDocument{%
  }

\setcopyright{acmcopyright}
\copyrightyear{2018}
\acmYear{2018}
\acmDOI{XXXXXXX.XXXXXXX}

\acmConference[Conference acronym 'XX]{Make sure to enter the correct
  conference title from your rights confirmation emai}{June 03--05,
  2018}{Woodstock, NY}
\acmPrice{15.00}
\acmISBN{978-1-4503-XXXX-X/18/06}

\usepackage{amsmath}
\usepackage{xspace}
\usepackage{algorithm}
\usepackage{algorithmic}
\usepackage{enumitem}

\makeatletter
\DeclareRobustCommand\onedot{\futurelet\@let@token\@onedot}
\def\@onedot{\ifx\@let@token.\else.\null\fi\xspace}

\def\eg{\emph{e.g}\onedot} 
\def\ie{\emph{i.e}\onedot}

\def\wrt{w.r.t\onedot} 
\def\aka{a.k.a\onedot}

\makeatother

\DeclareMathOperator*{\argmin}{arg\,min}

\newcommand{\tyger}{Tyger\xspace}

\definecolor{mycolor}{RGB}{219,90,107}

\begin{document}
\title[Tyger: Task-Type-Generic Active Learning for Molecular Property Prediction]{Tyger: Task-Type-Generic Active Learning \\for Molecular Property Prediction}

\author{Kuangqi Zhou}
\email{kzhou@u.nus.edu}
\affiliation{%
  \institution{National University of Singapore}
  \country{Singapore}
}

\author{Kaixin Wang}
\email{kaixin.wang@u.nus.edu}
\affiliation{%
  \institution{National University of Singapore}
  \country{Singapore}
}

\author{Jiashi Feng}
\email{jshfeng@gmail.com}
\affiliation{%
  \institution{ByteDance}
  \country{Singapore}
}

\author{Jian Tang}
\email{jian.tang@hec.ca}
\affiliation{%
  \institution{HEC Montreal, MILA}
  \country{Canada}
}

\author{Tingyang Xu}
\email{tingyangxu@tencent.com}
\affiliation{%
  \institution{Tencent}
  \country{China}
}

\author{Xinchao Wang}
\email{xinchao@nus.edu.sg}
\affiliation{%
  \institution{National University of Singapore}
  \country{Singapore}
}

\begin{abstract}
How to accurately predict the properties of molecules is an essential problem in AI-driven drug discovery, which generally requires a large amount of annotation for training deep learning models.
Annotating molecules, however, is quite costly because it requires lab experiments conducted by experts.
To reduce annotation cost, deep Active Learning~(AL) methods are developed to select only the most representative and informative data for annotating.
However, existing best deep AL methods are mostly developed for a single type of learning task (\eg, single-label or multi-label classification), and hence may not perform well in molecular property prediction that involves various task types.
In this paper, we propose a \textbf{T}ask-t\textbf{y}pe-\textbf{ge}ne\textbf{r}ic active learning framework (termed \textbf{Tyger}) that is able to handle different types of learning tasks in a unified manner.
The key is to learn a chemically-meaningful embedding space and perform active selection fully based on the embeddings, instead of relying on task-type-specific heuristics (\eg, class-wise prediction probability) as done in existing works.
Specifically, for learning the embedding space, we instantiate a querying module that learns to translate molecule graphs into corresponding SMILES strings (an expert-defined sequence molecular representation).
Furthermore, to ensure that samples selected from the space are both representative and informative, we propose to shape the embedding space by two learning objectives, one based on domain knowledge and the other leveraging feedback from the task learner (\ie, model that performs the learning task at hand).
We conduct extensive experiments on benchmark datasets of different task types.
Experimental results show that \tyger consistently achieves high AL performance on molecular property prediction, outperforming baselines by a large margin.
We also perform ablative experiments to verify the effectiveness of each component in \tyger.
\end{abstract}


\maketitle

\section{Introduction}
\label{sec: introduction}
Molecular property prediction is a fundamental step in computational drug discovery.
For accurately predicting molecular properties, deep learning methods have been widely applied, and have achieved remarkable success~\cite{gilmer2017neural, hu2020strategies, wu2018moleculenet, yang2019analyzing}.
However, such success relies heavily on a large amount of annotation, which is particularly time-consuming and costly~\cite{gilmer2017neural, mayr2018large}, since annotating molecules generally requires lab experiments or complex theoretical computation.

One promising way to alleviate this problem is Active Learning~(AL), which aims to find a strategy for selecting samples that are both \textbf{representative} in the data space and \textbf{informative} to the model, thus maximizing model performance under a low annotation budget.
In designing AL methods for molecular property prediction, since the molecular properties are of various types (\eg, binary-valued toxicity~\cite{richard2016toxcast}, real-valued electron energy gap~\cite{ramakrishnan2014quantum}), we need to accordingly consider different task types, such as single-label classification, multi-label classification, and regression.
Therefore, an AL method that can be generically applied to various types of tasks (which we term as ``task-type-generic'') would be desired.

However, most existing state-of-the-art deep AL methods are developed for a single type of learning task (usually single-label classification~\cite{ash2019deep,kirsch2019batchbald,sener2018active,choi2021vab}), and thus they may perform poorly on, or even cannot be applied to other task types (\eg, regression).
For example, \cite{kirsch2019batchbald, choi2021vab} rely on the model's class-wise prediction probability to select samples, which is unique to the classification task and cannot be extended to regression (more discussion are in Sec.~\ref{sec: related works}).

In this paper, we propose \textbf{Tyger}: a \textbf{T}ask-t\textbf{y}pe-\textbf{ge}ne\textbf{r}ic active learning framework for molecular property prediction.
The key to achieve task-type generality is to perform active selection on a learned molecular \emph{embedding space}.
In doing so, the selection process is only based on the embeddings, while not relying on heuristics unique to the task type at hand (\eg, aforementioned class-wise prediction for classification).
Specifically, for learning the embedding space, \tyger instantiates a \emph{querying module} that encodes molecule graphs into embeddings, and then decodes the embeddings back into the SMILES strings (\ie, a sequence molecular representation).
This allows the querying module to model correspondences between molecule substructures (\eg, functional groups) and substrings in SMILES, and thus the learned embedding space is chemically meaningful.

Furthermore, since the active selection is based on the embeddings, we want the embedding space to encode information about data \textbf{representativeness} and \textbf{informativeness} \wrt the task learner, which are two main sampling criteria for AL. To achieve this, we propose to shape the space with two simple yet effective training objectives in the following.

The first objective shapes the embedding space by pulling together embeddings of chemically-similar molecules, while pushing away those of dissimilar ones.
In this way, this objective encourages the embedding space to preserve global domain knowledge about the whole molecule data space (thus we refer to it as knowledge-aware objective). 
Such knowledge is valuable for for ensuring representativeness of selected data.
For example, some molecules are structurally similar (\eg, sharing same functional groups) and thus should be embedded closely.
Without such domain knowledge, even if we select a diverse set of embeddings from the space, the corresponding molecules may however lack chemical diversity, and are thus not representative for the data space.

The second objective, which we term as task-feedback objective, leverages feedback from the task learner to guide the learning of the embedding space.
Particularly, it distills information of hidden representations from the task learner into the embedding space.
The motivation behind this objective is that the informative samples (\eg, out-of-distribution ones~\cite{xie2021active}) often have special patterns in their hidden representations~\cite{lee2018simple}.
Therefore, distilling the hidden representations transfers such useful patterns to the embedding space.

The above two objectives endow the embedding space with two desirable properties respectively: (1) molecules whose embeddings are scattered in the space are also chemically diverse and hence form a representative subset of the molecule space; (2) informative molecules show special patterns in their embeddings. With these properties, we can pick representative and informative samples by simply choosing the samples whose embeddings are most \emph{dissimilar} to those of the labeled ones.
Specifically, when learning the embedding space, the querying module is also adversarially trained to distinguish between embeddings of labeled and unlabeled molecules.
Once training is finished, the querying module is able to identify whether a sample is similar to the labeled pool.

To evaluate our \tyger, we train a Graph Neural Network~(GNN) on various molecular property prediction datasets of different machine learning tasks~\cite{wu2018moleculenet}. 
Extensive results show that \tyger helps GNN to achieve high performance with limited training budget, and to outperform other competitive baseline AL methods by a large margin.
We also conduct ablation experiments to verify the effectiveness of different components of \tyger, and experimentally analyze how the proposed knowledge-aware and task-feedback objective contributes to the AL performance of \tyger.

In summary, our contribution are three-fold:
\begin{itemize}[leftmargin=20pt]
\item We propose \tyger, an active learning framework for molecular property prediction, which is generically applicable to various types of learning tasks in a unified manner.
\item In our proposed \tyger, we consider both global domain knowledge about the whole molecule space, and information feedback from the task learner . Thus, it could selected samples that are both representative and informative.
\item We demonstrate through extensive experiments that the \tyger enables GNNs to achieve high performance with low annotation budget on molecular property prediction.
\end{itemize}

\section{Related works}
\label{sec: related works}
\subsection{Molecular property prediction} 
Molecular property prediction is a critical step for computational drug discovery~\cite{sliwoski2014computational, yang2019analyzing}, and involves various machine learning tasks~\cite{wu2018moleculenet}.
Traditional methods (\eg, based on density functional theory~\cite{hohenberg1964inhomogeneous}) are too slow to be applied in practical scenarios~\cite{gilmer2017neural}, where generally a large number of molecules are to be processed.
To address this issue, deep learning methods~\cite{gilmer2017neural, hu2020strategies, yang2019analyzing, schutt2017schnetv} are proposed and widely adopted.
Among them, approaches based on (Graph Neural Networks)~GNNs have been shown to be very promising~\cite{gilmer2017neural, schutt2017schnetv, hu2020strategies, yang2019analyzing, klicpera2019directional}.
In particular, \citet{gilmer2017neural} propose the message passing neural network that is able to accurately predict various molecular properties. 
\citet{hu2020strategies} propose the GINE architecture by adapting the GIN architecture~\cite{xu2018powerful} and incorporating edge (\ie, chemical bond) information in each layer.
Due to its efficiency and high performance, GINE is widely adopted in works on molecular property prediction~\cite{hu2020strategies, guo2021few, wang2021property, zhang2021motif}.
In this work, we thus focus on GNN-based molecular property prediction.

\subsection{Active learning}
Active learning is a promising approach to alleviate the data-hunger issue in deep learning~\cite{settles2009active, ren2021survey}.
AL methods select (\aka query~\cite{settles2009active}) data samples according to two criteria: representativeness \wrt the whole data space and informativeness \wrt the task learner.
Here we review recent high-performance AL methods for deep models, and briefly discuss the difficulties of employing them to molecular property prediction.

Some AL works rely on the task learner to perform active selection, \eg, by using heuristic learner-based sampling rules ~\cite{sener2018active, kirsch2019batchbald, choi2021vab} or performing clustering~\cite{ash2019deep} on hidden representations of task learner.
These methods are developed based on the premise that the machine learning task of interest is single-label classification.
Therefore, when applied to multi-label classification or regression (if applicable), the performance would be inferior, since such extension might violate some key theoretical assumptions~\cite{sener2018active, ash2019deep}.
Moreover, many of these methods~\cite{ kirsch2019batchbald, choi2021vab, tan2021diversity} cannot be extended to regression, since they need class-wise prediction distribution or representation centers.

Towards being task-type-generic, one line of works~\cite{sinha2019variational, kim2021task, zhang2020state, mottaghi2019adversarial, wang2020dual} employ a querying module to select samples without making assumption of the learning task at hand.
For example, \citet{sinha2019variational} design a variational adversarial querying module by combining VAE~\cite{kingma2013auto} and GAN~\cite{goodfellow2014generative}, which is widely adopted and improved in subsequent works~\cite{kim2021task, zhang2020state, mottaghi2019adversarial, wang2020dual}.
However, there are two barriers blocking those works to be directly applied to GNN-based molecular property prediction.
First, they need to train their querying module is trained via input reconstruction, which is highly challenging for molecules, due to graph isomorphism~\cite{lim2018molecular, dollar2021attention, gomez2018automatic}.
Though some works can partly relieve this issue~\cite{kwon2019efficient, liu2018constrained, jin2018junction}, they suffer from one or more of the following drawbacks: inexact reconstruction~\cite{kwon2019efficient}, reliance on manually-defined graphlets~\cite{jin2018junction, jin2020hierarchical} or rules~\cite{liu2018constrained} for ensuring chemical validity of reconstructed molecules, and failure to reconstruct large molecules (as reported in \cite{jin2020hierarchical}).
Second, domain knowledge of molecule data space are not considered, and thus the selected training data of these methods might lack chemical diversity, which hurts task learner performance~\cite{zhang2019bayesian}.

Recently, most AL algorithms focus on image data. 
However, molecular data may need more helps from AL than image data because annotating molecules generally needs costly lab experiments conducted by experts. 
To the best of our knowledge, there is only one work, called ASGN~\cite{hao2020asgn}, that investigates molecule-targeted AL.
However, ASGN considers the semi-supervised active learning setting.
The technical contribution of ASGN is to combine their designed molecule-specific Semi-Supervised Learning~(SSL) tasks with an AL component based on CoreSet~\cite{sener2018active}, which is not task-type-generic. 
By contrast, we focus on the designing new AL algorithms for molecules, and our \tyger can collaborate with arbitrary SSL approaches.

\section{Notations and problem settings}
\label{sec: notations and settings}
Denote a molecule graph as $G=(V,E)$, where $V$ is the set of nodes (atoms), and $E=\{(u, v)|u, v \in V\}$ is the set of edges (chemical bonds).
Let $\mathbf{x}_u$ and $\mathbf{e}_{uv}$ denote the feature vector of node $v$ and edge $(u,v)$ respectively.
We may be interested in $n$ different properties of a molecule (\eg, toxicity and solubility), which are denoted by an $n$-dimension label vector $\mathbf{y}$ whose entries can be continuous or discrete.
In addition, let $S$ denote the SMILES string~\cite{weininger1988smiles} of a molecule graph $G$, which can be easily pre-generated by open cheminformatics libraries.

We consider batch-mode pool-based active learning~\cite{settles2009active}, a practical and widely-studied AL setting for deep models~\cite{sener2018active, sinha2019variational, kirsch2019batchbald, ash2019deep}. 
In this setting, we are given an initial labeled molecule pool $\mathcal{D}_L^0 = \{(G_i, \mathbf{y}_i)\}_{i=1}^{N_L^0}$, and a much larger unlabeled pool $\mathcal{D}_U^0 = \{G_i\}_{i=1}^{N_U^0}$. 
Our goal is to design an AL algorithm that performs $T$ rounds of \emph{query}~\cite{settles2009active}, \ie, sample selection. 
In the $t$-th round $(1 \leqslant t \leqslant T)$, a batch of $b$ samples, denoted as $\mathcal{B}^t$, are selected from $\mathcal{D}_U^{t-1}$.
Then, the selected samples are annotated by an oracle (\eg, a chemist), and moved from the unlabeled pool to the labeled one. 
Formally, let $\mathcal{B}^t_\textrm{anno} = \{(G_i, \mathbf{y_i}) | G_i \in \mathcal{B}^t\}$ denote annotated selected batch, then the pools are updated by $\mathcal{D}_L^t = \mathcal{D}_L^{t-1} \cup \mathcal{B}_\textrm{anno}^t$, and $\mathcal{D}_U^t = \mathcal{D}_U^{t-1} \backslash \mathcal{B}^t$; accordingly, $N_L^t=N_L^{t-1} + b$, $N_U^t=N_U^{t-1} - b$.
The obtained $\mathcal{D}_L^t$ is then used to train a GNN, \ie, the task learner. 

Note that, the union of $\mathcal{D}_L^t$ and $\mathcal{D}_U^t$ is the whole training set, \ie,  $\mathcal{D}_{\textrm{train}} = \mathcal{D}_L^t \cup \mathcal{D}_U^t, \forall t \in \{0,\cdots,T\}$.
In addition to $\mathcal{D}_{\textrm{train}}$, we also have a validation set $\mathcal{D}_\textrm{val}$ and a test set $\mathcal{D}_\textrm{test}$, which are held-out and disjoint from $\mathcal{D}_\textrm{train}$, for performing model selection and test on the task learner. 

For simplicity, we omit the round index $t$ unless necessary in the following part of this paper.

\section{Method}
\label{sec: method}

\begin{figure}[t!]
    \centering
    \includegraphics[width=0.95\linewidth]{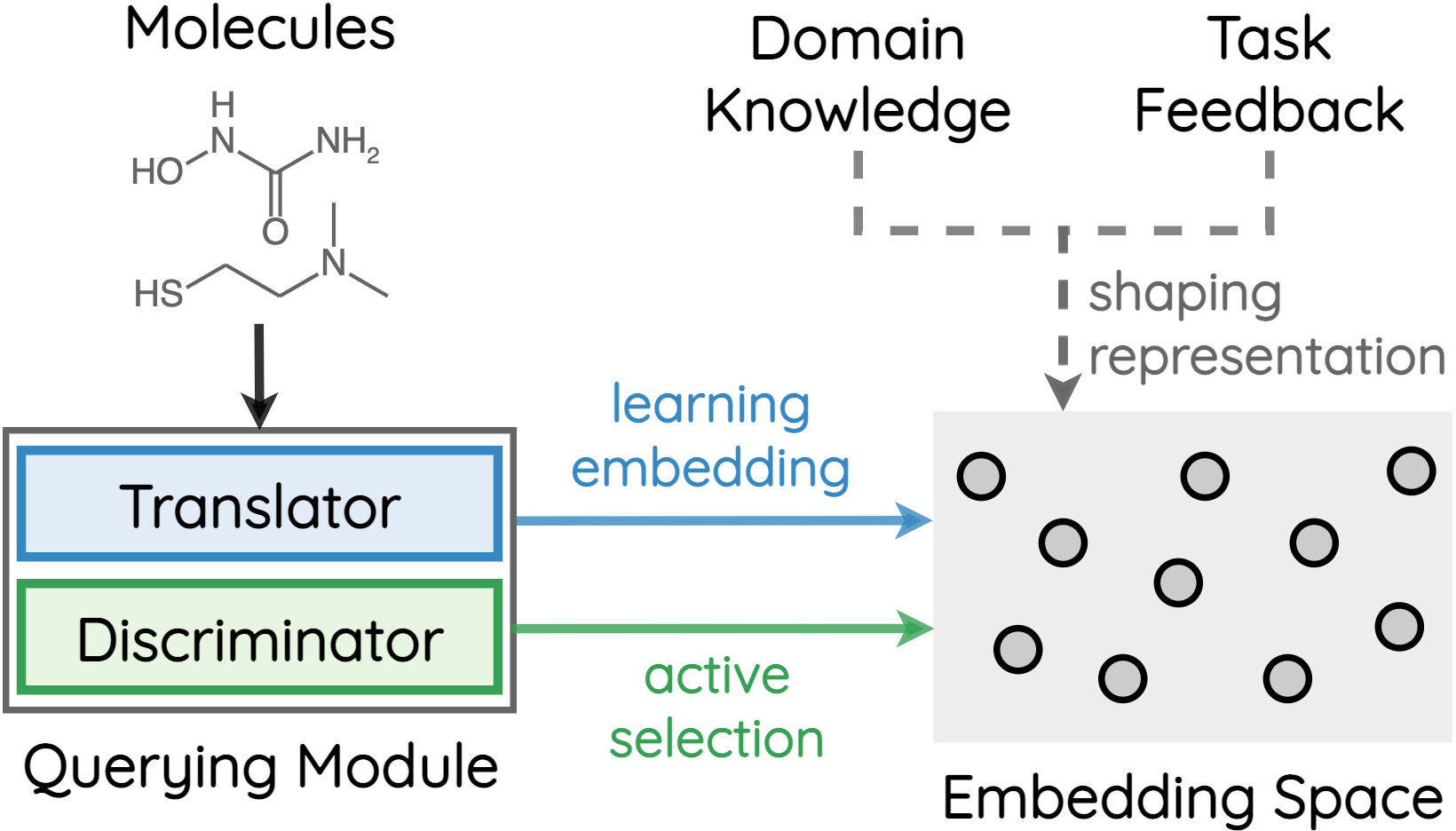}
    \caption{Overview of our proposed \tyger, a task-type-generic active learning framework that is able to handle different types of learning task in a unified manner.}
    \label{fig: overview}
\end{figure}

\begin{figure*}[t!]
    \centering
    \includegraphics[width=\linewidth]{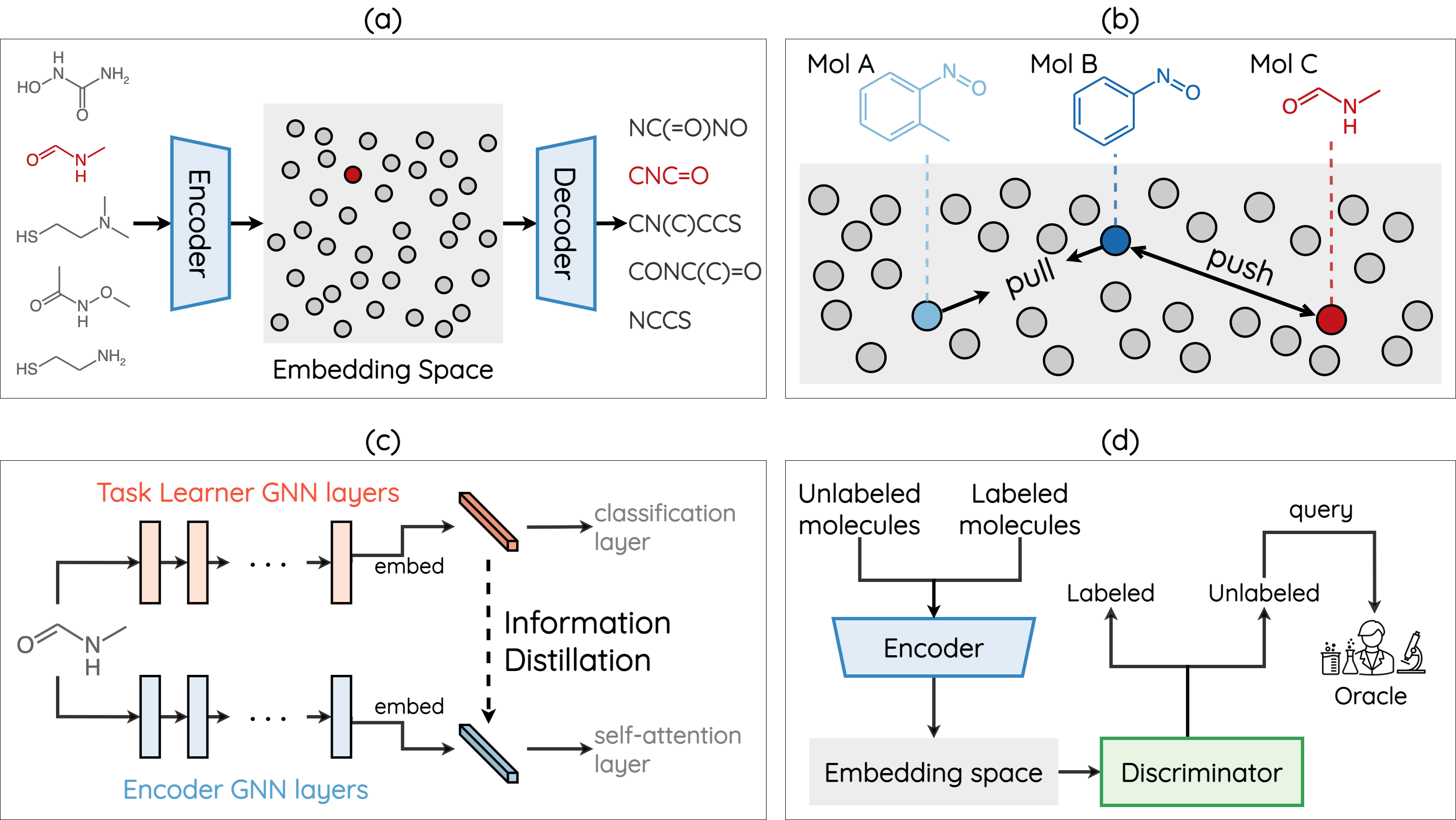}
    \caption{
    (a) The molecule translation process. 
    (b) Illustration of the knowledge-aware objective. 
    (c) Illustration of the task-feedback objective.
    (d) Illustration of the joint training of discriminator and the translator's encoder, and that of the adversarial-learning-based active selection strategy.}
    \label{fig: method}
\end{figure*}

We first provide an overview of the proposed \tyger. 
As illustrated in Fig.~\ref{fig: overview}, \tyger instantiates a \emph{querying module} to learn a chemically meaningful \emph{embedding space} used for active selection (Sec.~\ref{subsec: molecule translation}).
Furthermore, we leverage domain knowledge and feedback from the task learner (Sec.~\ref{subsec: information feedback}) to shape the embedding space (Sec.~\ref{subsec: domain knowledge}).
While learning the space, the querying module also learns a strategy for actively selecting samples in this space (Sec.~\ref{subsec: adversarial training}).
In doing so, the active selection process is fully based on the embeddings, making \tyger a task-type-generic method.

The querying module consists of two sub-networks: an encoder-decoder-like molecule translator (Sec.~\ref{subsec: molecule translation}) that learns the embedding space, and a discriminator (Sec.~\ref{subsec: adversarial training}) for active selection.

\subsection{Learning embedding space by molecule translation}
\label{subsec: molecule translation}
\tyger performs active selection on a learned embedding space. Therefore, the quality of the molecular embeddings is critical for AL performance. Prior works generally adopt an reconstruction reconstruction-based approach for learning the space, but such paradigm would not be suitable for molecule data due to the difficulty in reconstructing graphs (see Sec.~\ref{sec: related works} for more discussions).

We thus propose to learn the embedding space by training the querying module to translate molecules into their SMILES strings.
With this translation-based approach, the querying module is able to model correspondences between molecule substructures (\eg, functional groups) and substrings in SMILES, learning chemically meaningful embeddings without struggling to reconstruct them back to molecular graphs.

This is implemented by the molecular translator sub-network of the querying module, as illustrated in Fig.~\ref{fig: method}~(a)).

\textbf{The encoder} of the translator first applies $K$ message-passing-style~\cite{gilmer2017neural} GNN layers (\eg, the commonly-used GINE~\cite{hu2020strategies}) to the input graph $G_i$, and generate atom embeddings $\{\mathbf{h}_v^{(K)} |  v \in V\}$. Formally, the $k$-th GNN layer is:
\begin{equation}
\label{eqn: gnn layer}
    \begin{aligned}
    \mathbf{h}_v^{(k)} = \mathrm{U}^{(k)}\big(\mathbf{h}_v^{(k-1)}, \mathrm{M}(\{(\mathbf{h}_u^{(k-1)}, \mathbf{h}_v^{(k-1)}, \mathbf{e}_{uv}) | u \in \mathrm{N}(v)\})\big),
    \end{aligned}
\end{equation}
where $\mathbf{h}_v^{(k)}$ denotes atom $v$'s the hidden embedding in the $k$-th layer, $\mathrm{N}(v)$ is the 1-hop neighborhood of $v$, and $\mathrm{M}(\cdot)$, $\mathrm{U}(\cdot)$ denote the message and update functions.

After the GNN layers, the encoder applies a multi-head self-attention layer~\cite{vaswani2017attention} to capture long-range interactions among nodes, and then summarizes the node embeddings into a graph-level embedding $z$ via pooling:
\begin{equation}
\label{eqn: self attention}
    \begin{aligned}
    z = \mathrm{Pool}\big(\mathrm{Attn}(\{\mathbf{h}_v^{(K)} | v \in V\})\big),
    \end{aligned}
\end{equation}
where $\mathrm{Pool}(\cdot)$ and $\mathrm{Attn}(\cdot)$ denote the pooling and self-attention operations.

Long-range interactions among nodes are critical for learning on molecules, since atoms that are far from each other in the molecule graph might be very close in the real-world 3D space. 
Interactions among these atoms can shape molecule geometry through non-negligible forces such as electrostatic forces or van der Waals forces~\cite{luo2021predicting}, thus greatly influencing molecular properties.
However, the encoder might fail to capture such interactions if it applies GNN layers only --- each GNN layer only aggregates information of one hop of neighbors, while stacking many layers incur difficulty in GNN training~\cite{zhou2021understanding}.
We thus leverage the self-attention layer, which effectively models long-range dependencies in various tasks~\cite{vaswani2017attention, ying2021transformers}.

\textbf{The decoder} of the molecule translator then translates the molecule embeddings into their SMILES strings.
Since the decoding process is sequence modeling, we instantiate the decoder using a Transformer decoder~\cite{vaswani2017attention}.

In summary, the training objective of the molecular translator is:
\begin{equation}
\label{eqn: translation objective}
    \begin{aligned}
    L_{\textrm{trans}} = \mathbb{E}_{G_i \in \mathcal{D}_L \cup \mathcal{D}_U } \left[ \mathrm{SeqDist}(\mathrm{Dec}(\mathrm{Enc}(G_i)), S_i) \right],
    \end{aligned}
\end{equation}
where $\mathrm{Enc}(\cdot)$ and $\mathrm{Dec}(\cdot)$ are the encoder and decoder respectively, and $\mathrm{SeqDist}(\cdot, \cdot)$ denotes a distance between a pair of sequences. Particularly, We take the cross-entropy distance between tokenized sequences~\cite{vaswani2017attention, dollar2021attention}, since our decoder has a Transformer structure. 

\subsection{Improving representativeness by domain knowledge}
\label{subsec: domain knowledge}
For ensuring representativeness of the selected data, the learned embedding space should preserve global domain knowledge about the structure of the whole molecule space, \ie, chemical similarities among molecules.
For example, as shown in Fig.~\ref{fig: method}~(b), molecule B should be embedded closely to molecule A, since they they are structurally similar; by contrast, molecule B and molecule C should be far from each other in the embedding space.
However, such knowledge may not be gained by the molecule translator if it is only trained with the translation objective (Eqn.~\eqref{eqn: translation objective}).
The reason is that, when only trained to precisely model the relationship between individual graph-SMILES pairs, the translator mostly focuses on how local graph structures (\eg, functional groups) correspond to characters in SMILES substrings, while ignoring the relationship among molecules in the whole data space.

We thus propose to explicitly injects such domain knowledge into the embedding space, by shaping the space with a knowledge-aware objective inspired from Supervised Contrastive Learning (SCL) \cite{khosla2020supervised}. 
The core idea of SCL is to use some supervision (\eg, ground-truth class label in \cite{khosla2020supervised}) to define positive sets and negative sets used in contrastive learning.

In our case, the supervision is the desired global domain knowledge.
Specifically, we first model similarities among molecules (both labeled and unlabeled) according to some expert-defined molecular similarity metric (\eg, the widely-adopted Tanimoto similarity~\cite{bajusz2015tanimoto}), to compute an similarity matrix (\aka affinity matrix) $A \in \mathbb{R}^{(N_L + N_U) \times (N_L + N_U)}$, where $A_{ij}$ denotes the similarity between molecule $G_i$ and $G_j$.
Taking such similarities as a supervision, we define the positive and negative sets of a molecule $G_i$ as:
\begin{equation}
\label{eqn: positive and negative sets}
    \begin{aligned}
    \mathcal{P}_{i} &= \{G_j \in \mathcal{D}_L\cup\mathcal{D}_U| A_{ij} \geqslant \alpha\}, \\
    \mathcal{N}_{i} &= (\mathcal{D}_L\cup\mathcal{D}_U) \backslash \mathcal{P}_i.
    \end{aligned}
\end{equation}
where $\alpha$ is a threshold shared by all molecules.

Then, the knowledge-aware objective for shaping embedding space is:
\begin{equation}
\label{eqn: knowledge objective}
    \begin{aligned}
    L_\textrm{know} = \mathbb{E}_{G_i \in \mathcal{D}_L\cup\mathcal{D}_U}\! \left[\frac{-1}{|\mathcal{P}_{i}|}\!\sum_{G_j \in \mathcal{P}_{i}}\!\! \mathrm{log}\! \left(\frac{\mathrm{exp}(\hat{z}_i^\top\hat{z}_j)}{\sum_{G_k \in \mathcal{C}_{i}}\mathrm{exp}(\hat{z}_i^\top\hat{z}_k)}\right)\right],
    \end{aligned}
\end{equation}
where $\mathcal{C}_i = \mathcal{D}_L\cup\mathcal{D}_U \backslash \{G_i\}$, $\hat{z}_i = \frac{z_i}{\|z_i\|}$ is the normalized embedding of graph $G_i$, and $\tau$ is a temperature parameter.

As illustrated in Fig.~\ref{fig: method}~(b), this objective pulls together embeddings of molecules that have high chemical similarity, while pushing away dissimilar ones, thus encouraging the learned embedding space to preserve desired domain knowledge.
This helps the \tyger to select representative samples and achieve better AL performance (as empirically shown in Sec.~\ref{subsec: ablation} and Sec.~\ref{subsec: experimental analysis}).

\subsection{Enhancing informativeness by task learner feedback}
\label{subsec: information feedback}
Informativeness \wrt the task learner is also a main criterion in deep AL. For enhancing informativeness of the selected data, we design a feedback mechanism that distills information from the task learner to the querying module, as illustrated in Fig.~\ref{fig: method}~(c).

In particular, we propose an task-feedback objective that encourages the translator encoder's graph-level representation outputted by its GNN layers to be similar to that of the task learner.
This is inspired by observations that out-of-distribution samples~\cite{xie2021active} and adversarial samples~\cite{ducoffe2018adversarial} may be informative to the task learner, and that these samples often have special patterns in their hidden representations~\cite{lee2018simple}. 
Our feedback objective can transfer such useful patterns to the embedding space.

Formally, assume we distill information from the $k$-th layer of the task learner into the corresponding GNN layer of the encoder. 
Let $H^{(k)}_\textrm{enc}$ and $H^{(k)}_\textrm{task}$ denote the sets of atom embeddings (of the same input graph $G_i$) of $k$-th GNN layer of the encoder and task learner. 
We first apply a pooling operation to the atom embedding sets for generating graph-level representations $\mathbf{g}^{(k)}_\textrm{enc}$ and $\mathbf{g}^{(k)}_\textrm{task}$.
Then, the task-feedback objective is:
\begin{equation}
\label{eqn: feedback objective}
\begin{aligned}
L_\textrm{feed} = \mathbb{E}_{G_i \in \mathcal{D}_L \cup \mathcal{D}_U} \big[\mathrm{CosDist}(\mathbf{g}^{(k)}_\textrm{enc}, \mathbf{g}^{(k)}_\textrm{task}) \big],
\end{aligned}
\end{equation}
where $\mathrm{CosDist}(\cdot, \cdot)$ denotes the cosine distance between a pair of embeddings. 

This objective is used to train the translator only (\ie, the task learner is fixed when training the querying module), and works together with the translation and knowledge-aware objectives (Eqn.~\eqref{eqn: translation objective} and Eqn.~\eqref{eqn: knowledge objective}). 

We empirically show in Sec.~\ref{subsec: ablation} and Sec.~\ref{subsec: experimental analysis} that the task-feedback objective indeed encourages the \tyger to select samples that are more informative to the task learner and hence contributes to better AL performance.
  
\subsection{Learning active selection by adversarial training}
\label{subsec: adversarial training}
The \tyger performs active learning on the embedding space, by identifying and selecting molecules whose embeddings are most dissimilar to those of the current labeled ones.
Here we would like to highlight that, by doing so we can select data that are not only representative, but also informative.
This is because, with the help of the knowledge-aware objective (Eqn.~\eqref{eqn: knowledge objective}) and the task-feedback objective (Eqn.~\eqref{eqn: feedback objective}), the embedding space has the following two desirable properties:
\begin{itemize}[leftmargin=10pt, topsep=0pt]
 \setlength\itemsep{0em}
    \item Embeddings scattered in the embedding space are also diverse in the true molecule space.
    \item Embeddings of informative samples show some different patterns from those of uninformative ones (\eg, labeled samples).
\end{itemize}
Therefore, by selecting molecules dissimilar to the labeled ones in the embedding space, we can obtain a labeled pool that are both representative and informative.

In particular, when the querying module learns the embedding space, we also adversarially train it to distinguish between embeddings of labeled and unlabeled samples.
As illustrated in Fig.~\ref{fig: method}~(d), this is implemented by a discriminator sub-network of the querying module.
The discriminator is built upon the embedding space, and is jointly trained with the translator encoder in an adversarial manner --- the discriminator tries to distinguish the embeddings, while the encoder aims to fool the discriminator.
Formally, adversarial objectives for the discriminator (denoted as $\mathrm{D}(\cdot)$) and the encoder are:
\begin{equation}
\label{eqn: discriminator objective}
    \begin{aligned}
    L_{\text{D}} = -\mathbb{E}_{G \in \mathcal{D}_L} \big[\mathrm{log}(\mathrm{D}(z))\big] -\mathbb{E}_{G \in \mathcal{D}_U} \big[1-\mathrm{log}(\mathrm{D}(z))\big],
    \end{aligned}
\end{equation}
\begin{equation}
\label{eqn: translator adversarial objective}
    \begin{aligned}
    L_{\textrm{adv}} = -\mathbb{E}_{G \in \mathcal{D}_L} \big[\mathrm{log}(\mathrm{D}(z))\big] -\mathbb{E}_{G \in \mathcal{D}_U} \big[\mathrm{log}(\mathrm{D}(z))\big].
    \end{aligned}
\end{equation}

\begin{algorithm}[t!]
  \caption{Active learning performed by \tyger}
  \label{alg: tyger}
  \begin{algorithmic}[1]
  \REQUIRE initial labeled pool $\mathcal{D}_L^0$, unlabeled pool $\mathcal{D}_U^0$, \#rounds $T$, \#queries per round $b$, training epochs $n$, querying module, task learner
  \FOR{$T$ rounds}
  \STATE \(\triangleright\triangleright\triangleright\) \textbf{train querying module} \(\triangleleft\triangleleft\triangleleft\)
    \FOR{$n$ epochs}
      \STATE update molecule translator with Eqn.~\eqref{eqn: full objective}
      \STATE update discriminator with Eqn.~\eqref{eqn: discriminator objective}
    \ENDFOR
    \STATE \(\triangleright\triangleright\triangleright\) \textbf{perform active selection} \(\triangleleft\triangleleft\triangleleft\)
    \STATE select a batch of $b$ samples $\mathcal{B}^t$ according to Eqn.~\eqref{eqn: query samples}
    \STATE annotate $\mathcal{B}^t$ and obtain 
    $\mathcal{B}^t_\textrm{anno}$
    \STATE  $\mathcal{D}_L^t = \mathcal{D}_L^{t-1} \cup \mathcal{B}_\textrm{anno}^t$
    \STATE $\mathcal{D}_U^t = \mathcal{D}_U^{t-1} \backslash \mathcal{B}^t$
    \STATE \(\triangleright\triangleright\triangleright\) \textbf{train and test task learner} \(\triangleleft\triangleleft\triangleleft\)
    \STATE train task learner with $\mathcal{D}_L^t$, perform model selection using $\mathcal{D}_\textrm{val}$, and test the learner on $\mathcal{D}_\textrm{test}$
    \ENDFOR
  \ENSURE task learner performance on $\mathcal{D}_\textrm{test}$ in each round
  \end{algorithmic}
\end{algorithm}

Now we obtain train full training objective (denoted as $L_\textrm{full}$) of the molecule translator:
\begin{equation}
\label{eqn: full objective}
    \begin{aligned}
    L_\textrm{full} 
    =  L_\textrm{trans} + L_\textrm{adv} + \lambda_1 L_\textrm{know} + \lambda_2 L_\textrm{feed}.
    \end{aligned}
\end{equation}
We only introduce tunable weights (\ie, $\lambda_1$ and $\lambda_2$) for $L_\textrm{know}$ and $L_\textrm{feed}$
since we find in our pilot experiments that simply fixing the weights for $L_\textrm{trans}$, $L_\textrm{adv}$ to be 1 is sufficient for good performance. 
This also reduces the number of hyperparameters and hence facilitates model selection. 


Then, as can be seen from Eqn.~\eqref{eqn: discriminator objective}, a smaller output score $\mathrm{D}(z)$ of means that the corresponding input molecule are believed to have a larger probability of being unlabeled, and thus are more dissimilar to the labeled pool.
Therefore, once training is done, we use the querying module to select a batch of queries $\mathcal{B}=(G_1, \cdots, G_b)$ with the following strategy (see Fig.~\ref{fig: method}~(d)):
\begin{equation}
\label{eqn: query samples}
    \begin{aligned}
    \argmin\limits_{\mathcal{B} \subseteq \mathcal{D}_U}  (\mathrm{D}(G_i), \cdots, \mathrm{D}(G_b))
    \end{aligned}
\end{equation}

Alg.~\ref{alg: tyger} summarizes the active learning procedure of \tyger.
  
\section{Experiments}
\label{sec: experiments}
In this section, we evaluate our \tyger with experiments. 
We first describe experiment settings in Sec.~\ref{subsec: exp settings}, and then compare the \tyger with competitive AL methods in Sec.~\ref{subsec: main exp}.
Furthermore, we conduct ablation studies to verify the effectiveness of different components of \tyger in Sec.~\ref{subsec: ablation}, and experimentally analyze on our method in Sec.~\ref{subsec: experimental analysis}.

\subsection{Experiment settings}
\label{subsec: exp settings}
We run experiments under the batch-mode pool-based active learning setting (see Sec.~\ref{sec: notations and settings} for details). 
We initialize the labeled pool by randomly selecting 10\% samples of the entire training set; the initial unlabeled pool is the rest 90\% of the training set.
Then, 10 rounds of active learning are performed.
In each round, an unlabeled batch of 4\% samples of the entire training set is actively selected, annotated, and moved from the unlabeled pool to the labeled one.
Thus, the total annotation budget is 50\% of the training set.

\noindent \textbf{Learning tasks and datasets.} 
We run experiments on both classification and regression tasks, using datasets from the commonly-adopted MoleculeNet benchmark~\cite{wu2018moleculenet}, which are also included in the Open Graph Benchmark (OGB)~\cite{hu2020open}.

\begin{figure}[t]
    \centering
    \includegraphics[width=\linewidth]{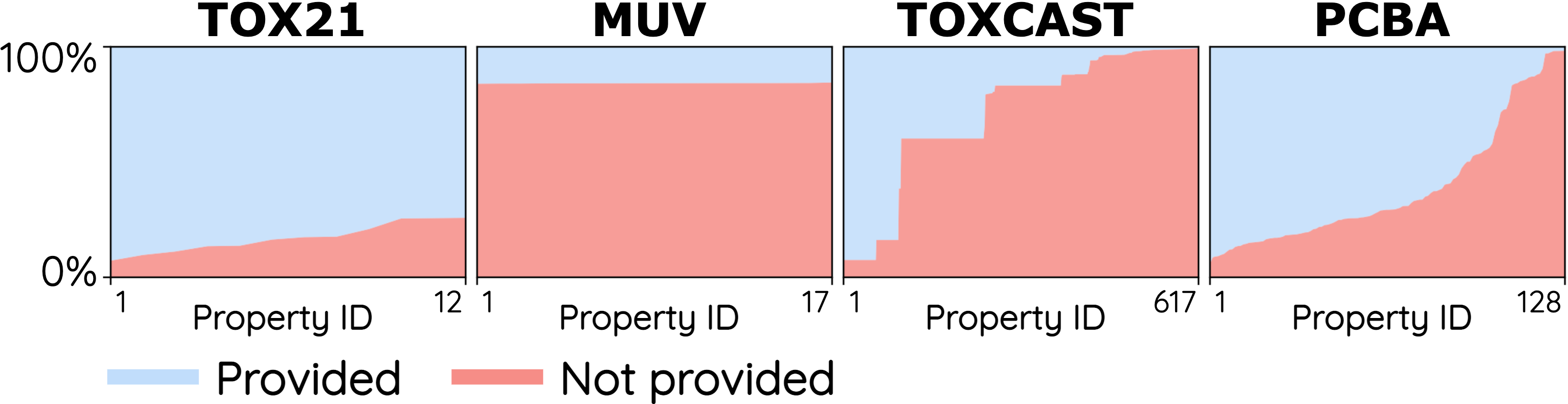}
    \caption{Ratios of molecules whose ground truth properties are provided or not. The x-axis ticks correspond to the index of properties. We sort the properties by the ratio of label-not-provided molecules in ascending order.}
    \label{fig: ratio}
\end{figure}

For (single-label or multi-label) classification, we use: 
\begin{itemize}[leftmargin=10pt, topsep=0pt]
 \setlength\itemsep{0em}
    \item BACE: binding results of a protein and its inhibitors;
    \item BBBP: the blood-brain barrier penetration property of molecules;
    \item HIV: measured results of ability to inhibit HIV replication;
    \item SIDER: side effects of marketed drugs. Note that SIDER includes results of 27 organs, and hence it is a multi-label dataset.
\end{itemize}
Though some other classification datasets in \cite{wu2018moleculenet} (\eg, TOX21 and PCBA) are also widely used, we find that they contain a large number molecules whose properties are not fully provided by the dataset creator (as shown in Fig.~\ref{fig: ratio}).
This violates the assumption of our setting that each queried sample is correctly annotated by an oracle, and thus we do not use these datasets. 

For regression, we use:
\begin{itemize}[leftmargin=10pt, topsep=0pt]
 \setlength\itemsep{0em}
    \item ESOL: water solubility of molecules;
    \item LIPOPHILICITY: measured results of octanol-water distribution coefficient, a commonly-used measurement of molecules' lipophilicity; We use ``LIPO'' to abbreviate ``LIPOPHILICITY''.
\end{itemize}

Following \cite{hu2020strategies}, we use \textit{scaffold split}, which has been shown to provide a more realistic estimate of model performance than random split.
The split for training/validation/test sets is 80\%/10\%/10\%. Statistics of used datasets are in Tab.~\ref{tab: dataset statistics}.

\begin{table}[t]
  \caption{Statistics of used datasets}
  \label{tab: dataset statistics}
  \begin{tabular}{llll}
    \toprule
    Dataset & Task type & \# Labels & \# Molecules\\
    \midrule
    BACE & classification & 1 & 1,513 \\
    BBBP & classification & 1 & 2,039 \\
    SIDER & classification & 27 & 1,427 \\
    HIV & classification & 1 & 41,127 \\
    ESOL & regression & 1 & 1,128 \\
    LIPO & regression & 1 & 4,200 \\
  \bottomrule
\end{tabular}
\end{table}

\noindent \textbf{Implementation details.}
We instantiate the task learner and the GNN part of our translator's encoder with a 5-layer GINE architecture~\cite{hu2020strategies}, which is widely used for molecular property prediction~\cite{hu2020strategies, guo2021few, wang2021property, zhang2021motif}. 
The translator's decoder is a Transformer decoder. 
The discriminator of the querying module consists of a self-attention layer, a 2-layer MLP and a Sigmoid activation.
We use the RdKit library\footnote{https://www.rdkit.org/} to pre-generate SMILES.
For the chemical similarity used in the knowledge-aware objective (Sec.~\ref{subsec: domain knowledge}), 
we use Tanimoto similarity~\cite{bajusz2015tanimoto} to build the similarity matrix $A$, which can be easily calculated via RdKit.

In each AL round, the molecule translator and the discriminator are iteratively trained for 50 epochs, with batch size of 128 and learning rate of 5e-4.
The task learner is trained for 50 epochs with batch size of 128 and learning rate of 1e-3.
The used optimizer is Adam.
Note that the task learner is re-initialized before it is trained and tested.

The thresholds $\alpha$ for obtaining positive sets (see Eqn.~\eqref{eqn: positive and negative sets}) are set to be the 2-nd quantile (for BACE and BBBP) or 3-rd quantile (for other datasets) of all pair-wise similarities $\{A_{ij}|i \geqslant j\}$.
For the feedback mechanism (Sec.~\ref{subsec: information feedback}), we distill the information from the 3-rd layer (for BACE and ESOL) or 4-th layer (for other datasets).
We perform hyperparameter search for tuning $\lambda_1$ and $\lambda_2$.

\begin{figure*}[t]
    \centering
    \includegraphics[width=\linewidth]{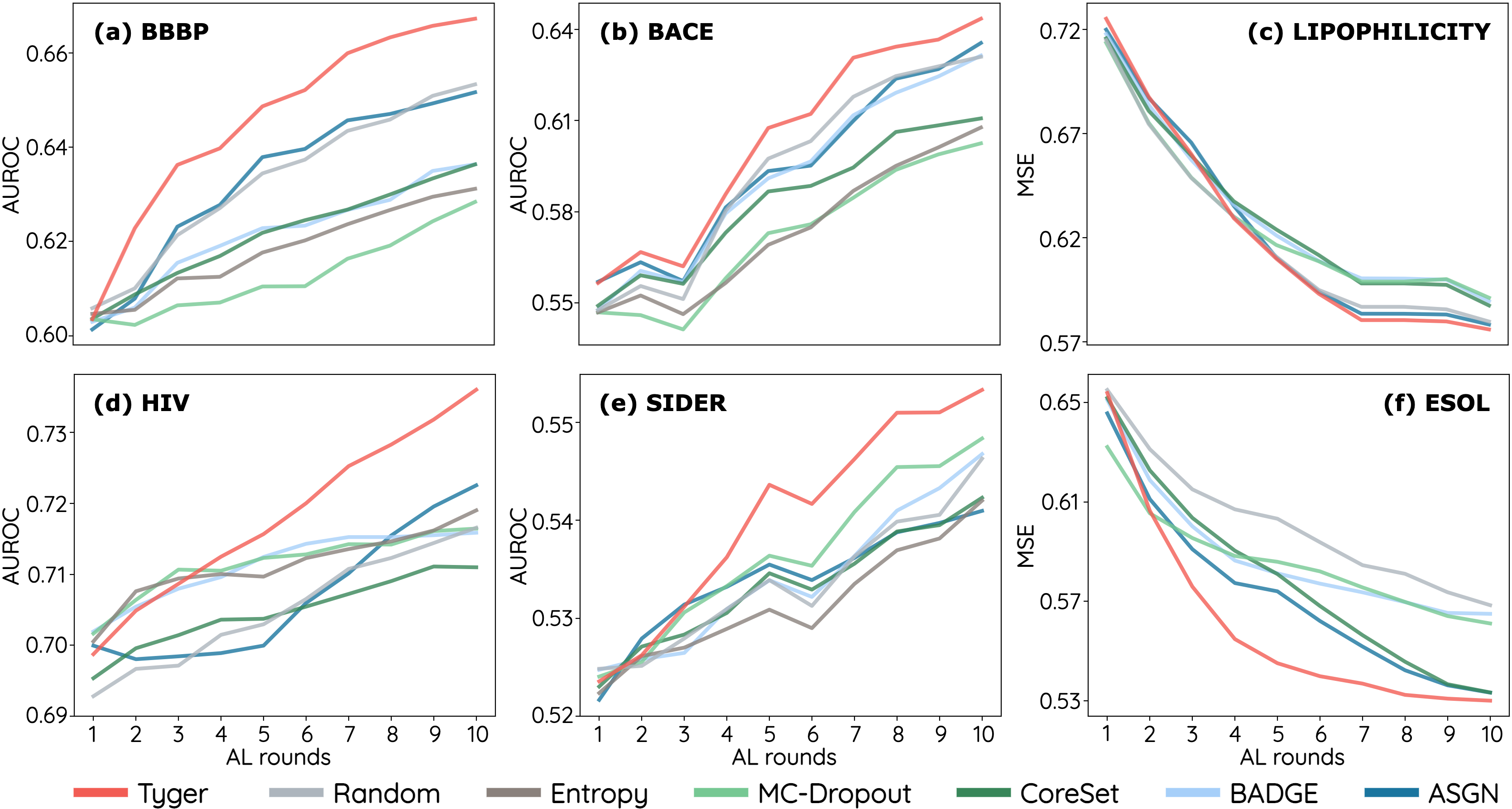}
    \caption{Active learning performance of different methods. Performance on initial labeled pools (\ie, round 0) is not shown, since all methods use the same initial pools.}
    \label{fig: main results}
\end{figure*}

\noindent \textbf{Performance evaluation.} 
For evaluation, we calculate performance of the task learner on the test set in each AL round.
The performance metrics are AUROC for classification and MSE for regression, respectively.
We randomly sample 50 initial labeled pools in advance, and run each AL experiment with all these pools and 10 random seeds per pool.
The reported performance is averaged across all initial pools and seeds.

\subsection{Active learning performance}
\label{subsec: main exp}
We compare our proposed \tyger it with the following baselines:
\begin{itemize}[leftmargin=10pt, topsep=0pt]
\setlength\itemsep{0em}
\item Random: randomly selecting a batch of samples.
\item Entropy: selecting samples with the largest entropy of the task learner's softmax prediction distribution. 

For multi-label datasets, the entropy is averaged across all labels.
We do not run Entropy for regression datasets, where there is no softmax prediction.
\item MC-Dropout~\cite{gal2017deep}: estimating model uncertainty via Monte-Carlo dropout.
Uncertainty is averaged across all labels for multi-label datasets.
\item CoreSet~\cite{sener2018active}: selecting samples with largest minimum distance to the current training data.
\item{BADGE}~\cite{ash2019deep}: selecting samples by performing k-MEANS++ clustering on the loss function's gradients \wrt the weights of output layer of the task learner.
BADGE is the state-of-the-art method on many image classification datasets.
\item{ASGN}~\cite{hao2020asgn}: training a teacher-student framework and selecting samples using CoreSet~\cite{sener2018active} upon the representations of the teacher model.
To our best knowledge, ASGN is the only previous work that investigates molecule-targeted AL. 
\end{itemize}

\begin{figure*}[t!]
    \centering
    \includegraphics[width=\linewidth]{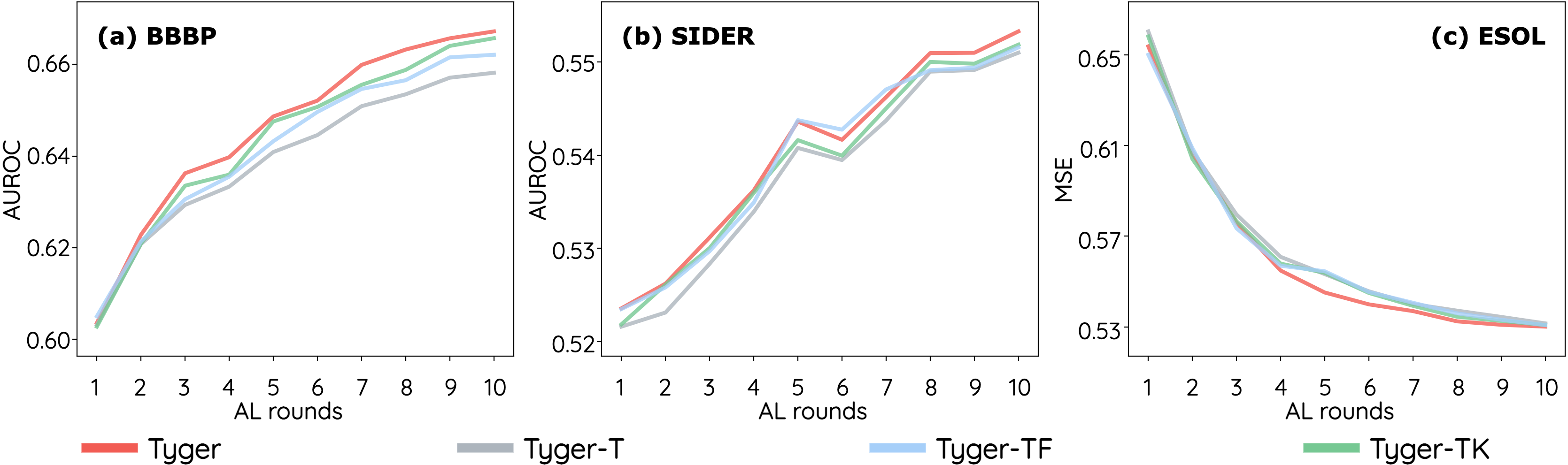}
    \caption{Active learning performance of \tyger and 3 variants of it: \tyger-T, \tyger-TK and \tyger-TF.}
     \label{fig: ablation}
\end{figure*}
\begin{figure*}[t!]
    \centering
    \includegraphics[width=\linewidth]{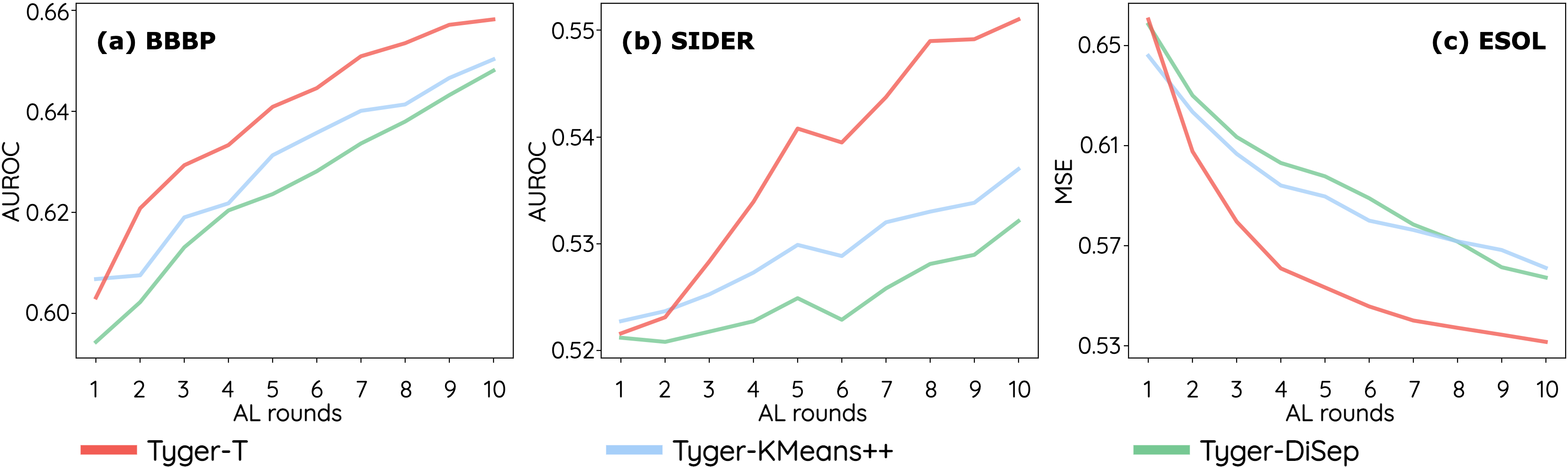}
    \caption{Active learning performance of \tyger-T, \tyger-KMeans++, and \tyger-DiSep.}
    \label{fig: adversarial}
\end{figure*}

We show results for all methods in Fig.~\ref{fig: main results}.
As can be seen, our \tyger achieves better performance than baselines on all 6 datasets. 
Specifically, on BBBP and HIV, \tyger outperforms the baselines by a large margin.
We attribute the consistently good performance to the fact that \tyger is a task-type-generic method.
By contrast, the methods developed for classification (\eg, BADGE), may perform unsatisfactorily on regression datasets.
Another key factor for the \tyger's good performance is that it considers both data representativeness and informativeness, which will be further discussed in Sec.~\ref{subsec: experimental analysis}.
In comparison, methods based on only representativeness or informativeness fail to achieve consistently good performance.
For example, MC-Dropout, based on informativeness only, performs well on SIDER but is the worst baseline on BBBP.

Additionally, we find that ASGN generally outperforms other baselines. 
In particular, though ASGN and CoreSet essentially use the same active selection strategy (\ie, largest minimum distance to current training data), ASGN outperforms CoreSet on all datasets. 
This might be because the semi-supervised learning tasks in ASGN introduce molecule-specific knowledge into its learned representations, which also supports our claim that domain-specific designs are critical for active learning on molecules.

\subsection{Ablation study}
\label{subsec: ablation}
We conduct experiments to study the effectiveness of the knowledge-aware and task-feedback objectives, and active selection strategy. 

\subsubsection{Effectiveness of knowledge-aware and task-feedback objectives}
\label{subsubsec: ablate objectives}
In Sec.~\ref{subsec: domain knowledge}, we propose an knowledge-aware objective (Eqn.~\eqref{eqn: knowledge objective}) to inject domain knowledge about the molecule space into the querying module's learned embedding space.
Furthermore, we propose in Sec.~\ref{subsec: information feedback} a task-feedback objective (Eqn.~\eqref{eqn: feedback objective}) that informs the querying module the on how its queried samples influence the task learner.
Here we ablatively study the effectiveness of these two objectives on AL.

To this end, we compare the performance of \tyger and 3 variants of it:
\begin{itemize}[leftmargin=10pt, topsep=0pt]

\item \textbf{Tyger-T}: training the querying module with the \textbf{T}ransla\-tion objective only (see Eqn.~\eqref{eqn: translation objective});

\item \textbf{Tyger-TK}: training the querying module with translation and \textbf{K}nowledge-aware objectives;

\item \textbf{Tyger-TF}: which is trained with translation and \textbf{F}eedback objectives.
\end{itemize}
All these 3 variants adopt the adversarial objectives (see Eqn.~\eqref{eqn: discriminator objective} and Eqn.~\eqref{eqn: translator adversarial objective}).

We run experiments on 3 datasets of different learning tasks: BBBP (single-label classification), SIDER (multi-label classification) and ESOL (regression). 
Other settings are the same as in Sec.~\ref{subsec: exp settings}.

As shown in Fig.~\ref{fig: ablation}, on BBBP and SIDER, \tyger-TK and \tyger-TF outperform \tyger-T, demonstrating that the two objectives contribute to the good performance of \tyger.
Furthermore, when they work together (which yields the original \tyger), they could further help achieve better performance.
Though \tyger-TK and \tyger-TF are not better than \tyger-T on ESOL, we would still like to highlight that the collaboration of these objectives is beneficial, which can seen from the fact that \tyger is better than \tyger-T also on this dataset.

\subsubsection{Effectiveness of active selection approach}
\label{subsubsec: ablate adversarial learning}
In our \tyger, we use an adversarial-learning-based approach to perform active selection on the embedding space of the querying module (Sec.~\ref{subsec: adversarial training}).
Here we conduct experiments to study the effectiveness of this approach.

Specifically, we compare the \tyger-T variant and another two variants that uses different selection strategies upon the embedding space:
\begin{itemize}[leftmargin=10pt, topsep=0pt]

\item \textbf{Tyger-KMeans++}: selecting samples by performing KMeans++ clustering on the embeddings and then choosing the centroids. It has been shown in \cite{ash2019deep} that the KMeans++ algorithm is able to selects a diverse set of samples and is hence a competitive selection approach.

\item \textbf{Tyger-DiSep}: instead of adversarially training the discriminator and the encoder, we first train the encoder to learn the embedding and fix it. Then, the  \textbf{Di}scriminator is \textbf{Sep}arately trained to distinguish between labeled and unlabeled embeddings. This can be seen as an implementation of \cite{gissin2019discriminative}.

\end{itemize}
Similar to \tyger-T, \tyger-KMeans++ or \tyger-DiSep are not trained with the knowledge-aware or task-feedback objective.

We run experiments on BBBP, SIDER and ESOL, using the settings specified in Sec.~\ref{subsec: exp settings}. As shown in Fig.~\ref{fig: adversarial}, \tyger-T consistently outperforms \tyger-KMeans++ and \tyger-DiSep, demonstrating the superiority of our used selection approach.

\subsection{Experimental analysis}
\label{subsec: experimental analysis}
As shown in Sec.~\ref{subsubsec: ablate objectives}, the knowledge-aware and task-feedback objectives contribute to the good AL performance of the \tyger.
In this subsection, we dig deeper into these two objectives by experimentally analyzing how they benefit the performance. Specifically, we investigate the following 3 questions:
\begin{itemize}[leftmargin=20pt, topsep=0pt]
 \setlength\itemsep{0em}
    \item[\textbf{Q1}:] whether the knowledge-aware objective encourages the querying module to preserve knowledge in the embedding space as expected;
    \item[\textbf{Q2}:] whether the knowledge-aware objective increases the representativeness of the selected data.
    \item[\textbf{Q3}:] whether the task-feedback objective indeed helps to select data that are informative to the task learner.
\end{itemize}

To study \textbf{Q1}, we compare \tyger-T and \tyger-TK by computing cosine similarities among their learned embeddings, and then calculating the ratio between the mean cosine similarity for all negative sets and that for all positive sets (defined in Eqn.~\eqref{eqn: positive and negative sets}). 
Formally, let $C_{ij}$ denote the cosine similarity between learned embeddings of $G_i$ and $G_j$, then the ratio is:
\begin{equation}
\label{eqn: pos-to-neg sim ratio}
\begin{aligned}
SimR = \frac{\sum\limits_{G_i \in \mathcal{D}_\textrm{train}} \frac{1}{|\mathcal{N}_i|} \sum\limits_{G_j \in \mathcal{N}_i} C_{ij}}{\sum\limits_{G_i \in \mathcal{D}_\textrm{train}} \frac{1}{|\mathcal{P}_i|} \sum\limits_{G_j \in \mathcal{P}_i} C_{ij}}.
\end{aligned}
\end{equation}
A smaller $SimR$ means that more knowledge is preserved --- positive pairs (\ie, molecules have high chemical similarity) are pulled closer, while negative pairs are pushed farther. 
For \textbf{Q2}, we compute the mean Tanimoto similarity (denoted as $mTaniSim$) among the samples selected by \tyger-T and those selected by \tyger-TK.
For \textbf{Q3}, we compute 4 metrics measuring informativeness for samples selected by \tyger-T and \tyger-TF: (i) entropy of softmax prediction ($Ent$), (ii) entropy estimated by MC-Dropout~\cite{gal2017deep} ($MC\textrm{-}Ent$), (iii) loss computed with the ground-truth label ($Loss$), as used in \cite{kim2021task}, and (iv) the norm of the loss's gradient \wrt weights of the classification layer ($GNorm$). 
As stated in \cite{ash2019deep}, a larger gradient norm implies that using the corresponding sample to train the task learner brings a larger change to the learner's weights, and thus brings more information to the task learner.

\begin{table}[t!]
  \caption{Metrics of data diversity of informativeness for 3 variants of \tyger. Results are on SIDER.}
  \label{tab: exp analysis}
  \begin{tabular}{lccc}
    \toprule
    Metric & \tyger-T & \tyger-TK & \tyger-TF \\
    \midrule
    $SimR \downarrow$ & 0.900 & 0.825 & -\\
    $mTaniSim \downarrow$ & 0.115 & 0.102 & - \\
    $Ent \uparrow$ & 0.518 & - & 0.532 \\
    $MC\textrm{-}Ent \uparrow$ & 0.553 & - & 0.560 \\
    $Loss \uparrow$ & 0.540 & - & 0.551 \\
    $GNorm \uparrow$ & 0.101 & - & 0.104\\
  \bottomrule
\end{tabular}
\end{table}

Tab.~\ref{tab: exp analysis} shows the results on SIDER.
We have the following observations.
\textbf{(1)} Embeddings of \tyger-TK has lower $SimR$ than those of \tyger-T, suggesting that the knowledge-aware objective indeed helps in injecting desired domain knowledge into the embedding space;
\textbf{(2)} The knowledge-aware objective helps to select samples that are more chemically diverse, and thus the selected training set is more representative in the whole data space (comparing the $mTaniSim$ of \tyger-T and \tyger-TK).
\textbf{(3)} The task-feedback objective is effective in encouraging the \tyger to select samples that are informative to the task learner, \eg, those with large model uncertainty and/or large prediction loss, and those that causes large change to the task learner. 

\section{Conclusion}
\label{sec: conclusion}
In this work, we propose a task-type-generic active learning framework for molecular property prediction (\tyger). 
Our \tyger performs active selection on a chemically meaningful embedding space learned by it, instead of relying on task-specific heuristics, which is the key to achieving task type generality.
For learning the space, we train an encoder-decoder-like querying module to model the correspondences between a molecule and its SMILES strings.
Furthermore, we introduce two novel learning objectives to guide the querying module learning.
The first objective encourages the module to preserve domain knowledge about the molecule space structure into the embedding space, while the second objective enables the querying module to receive feedback from the task learner.
With the two objectives, the querying module is able to select representative and informative samples, simply by identifying and choosing samples whose embeddings are most dissimilar to those of the labeled ones.
Experimental results on benchmark datasets of different learning tasks demonstrate the effectiveness of the \tyger.

\bibliographystyle{ACM-Reference-Format}
\bibliography{ref}

\end{document}